\documentclass{article}
\usepackage{spconf,amsmath,graphicx}
\usepackage{multirow}
\usepackage{color}
\usepackage{multirow}
\usepackage{booktabs}
\usepackage{caption} 
\usepackage{enumitem}
\usepackage[normalem]{ulem}
\newcommand{\comm}[1]{}
\title{Fusion approaches for emotion recognition from speech using acoustic and text-based features}

\name{Leonardo Pepino$^{\star}$ \qquad Pablo Riera$^{\star}$ \qquad Luciana Ferrer$^{\star}$ \qquad Agust\'in Gravano$^{\star \dagger}$}

\address{
  $^{\star}$Instituto de Investigaci\'on en Ciencias de la Computaci\'on (ICC), CONICET-UBA, Argentina\\
  $^{\dagger}$Departamento de Computaci\'on, FCEyN, Universidad de Buenos Aires (UBA), Argentina
  \thanks{This material is based upon work supported by the Air Force Office of Scientific Research under award number FA9550-18-1-0026.
  We gratefully acknowledge Carlos Busso for giving us access to MSP-PODCAST dataset, and NVIDIA Corporation for the donation of the Titan Xp GPU 
2
 used for this research.
  Correspondence: lpepino@dc.uba.ar}
}

\begin{document}
\ninept
\maketitle
\begin{abstract}
In this paper, we study different approaches for classifying emotions from speech using acoustic and text-based features. We propose to obtain contextualized word embeddings with BERT to represent the information contained in speech transcriptions and show that
this results in better performance than using Glove embeddings. We also propose and compare different strategies to combine the audio and text modalities, evaluating them on IEMOCAP and MSP-PODCAST datasets. We find that fusing acoustic and text-based systems is beneficial on both datasets, though only subtle differences are observed across the evaluated fusion approaches. Finally, for IEMOCAP, we show the large effect that the criteria used to define the cross-validation folds have on results. In particular, the standard way of creating folds for this dataset results in a highly optimistic estimation of performance for the text-based system, suggesting that some previous works may overestimate the advantage of incorporating transcriptions.
\end{abstract}

\vspace{-2mm}
\section{Introduction}
\label{sec:intro}

Speech emotion recognition (SER) is an active research area with important applications in the field of human-computer interaction. SER is a complex task even for humans \cite{devillers_challenges_2005}. In fact, in spite of recent advances enabled by deep learning models and the release of larger emotion datasets, the performance of SER systems is still relatively poor, with average recall rates usually well below 70\% on the most realistic datasets, indicating that it remains an open problem.

Most SER systems use low-level descriptors (LLD) extracted from the audio signal such as MFCCs, pitch and voice quality features \cite{anagnostopoulos_features_2015}, or features learned automatically from spectrograms using deep neural networks \cite{zheng_experimental_2015,satt_efficient_2017}. The excellent performance of current automatic speech recognition systems (ASR) also allows us to extract reliable text transcriptions from the speech without the need for human annotators. A few works have incorporated this information into SER systems. In some of these studies, emotional word-based vectors were computed from word occurrences in each emotion class \cite{jin-evector}, or using external lexicons \cite{chuang2004multi}. Similarly, emotional vectors can be extracted from phonemes \cite{gamage2017salience}. In some works \cite{jin-evector}, the word-based vectors are used as input to SVM classifiers together with high-level statistics of the acoustic LLDs. Another approach is to train text- and audio-based classifiers separately and combine their outputs to make a final prediction \cite{softfusion}. Recently, deep neural networks have been used to learn audio-linguistic embeddings \cite{haque2019audio} and to train emotion classifiers in an end-to-end framework combining text and audio modalities \cite{yoon2018multimodal,sebastian_fusion_2019,sahu_multi-modal_2019, zhang_exploiting_2019}.

In this paper, we study different ways of fusing audio and linguistic information, using early and late fusion techniques and comparing different training approaches, including (1) initializing two individual branches with models trained separately for audio and text and further fine-tuning the last few layers, (2) fixing the text and audio branches and training only the fusion parameters, and (3) training the whole combined neural network from scratch. For the audio branch, we use a standard approach based on MFCC, pitch, loudness, jitter, shimmer and logHNR features. For the text branch, we use contextualized word embeddings \cite{devlin2019bert} instead of the standard word embeddings like Glove \cite{glove} used in most of the previous works \cite{sahu_multi-modal_2019,yoon2018multimodal}. Standard word embeddings like those obtained with Glove are extracted independently of the context in which the words are found. For example, the word ``sad'' would be assigned the same embedding whether the phrase was ``I am very sad'' or ``I am not sad at all''. On the other hand, contextualized word embeddings like those extracted by BERT take into account the whole phrase in which the word is found. As a consequence, the embedding corresponding to the word ``sad'' in those two phrases would most likely be different. We hypothesized that this characteristic should positively impact SER performance. To our knowledge, \cite{sun_multi-modal_2019} is the only work in which word embeddings extracted with BERT have been used for SER. In that paper, authors propose a shared representation of audio, text and video modalities through deep canonical correlation analysis. A comparison with other types of embeddings is not shown in that work.

The proposed models are tested on the well-studied IEMOCAP dataset \cite{busso-IEMOCAP}, as well as on the more challenging MSP-PODCAST dataset \cite{busso-podcast}.  Our first contribution is to show that linguistic information gives significant improvements in performance when combined with acoustic information on the MSP-PODCAST dataset. As far as we know, this is the first time that linguistic information has been used on this dataset. Second, we show that the use of contextualized word embeddings obtained with BERT results in significant improvements with respect to using standard word embeddings obtained with Glove. Third, we propose a novel way to fuse audio and text information by pretraining the neural network in audio and text modalities and then fine-tuning the fused model. Finally, we show that creating folds by speaker is not sufficient to obtain fair performance predictions on IEMOCAP, since the data contains scripted dialogues which greatly affect the performance of text-based systems when the same script is observed in training and testing. This being such a widely used dataset, we believe this observation is of great importance to the research community.

\vspace{-2mm}
\section{Models}
\label{sec:models}

This section describes the models used in the experiments. First, the individual models for each modality are introduced. Then, models that combine the text and audio information are described. All models are trained to optimize cross-entropy loss for four emotion classes: happy, sad, angry and neutral.

\vspace{-2mm}
\subsection{Text-based model}

Recently, a language model called BERT \cite{devlin2019bert}, trained with large amounts of data has been released to the community. This model can be fine-tuned or used as a feature extractor for downstream tasks, achieving state-of-the-art results on many of them. BERT is based on the Transformer \cite{vaswani2017attention} -- a network capable of modeling long contextual information, generating word embeddings that are conditioned on the phrase in which the word is found. 
In this study, a sequence of word embeddings is extracted from speech transcriptions using the pretrained BERT base uncased model, which consists of 12 layers, 12 attention heads and 110M parameters. The word embeddings are formed by adding the activations of the last 4 layers of the pretrained BERT model without fine-tuning. The resulting features are used as input to the text model shown on the left of Figure~\ref{fig:branches}.

The first layer ($L_{T1}$) in our text-based model operates on each embedding in the sequence reducing its dimensionality from 768 to 128. Then, 2 convolutional layers ($L_{T2}$ and $L_{T3}$) model relationships across neighboring elements of the sequence. Finally, an average over time is taken, resulting in an embedding that summarizes all the information in the sample. A final dense layer with softmax activation predicts emotion probabilities $P(C_k)$. We applied batch normalization in all layers.

To make a comparison with non-contextualized word embeddings, we trained the same model using 300-dimensional Glove embeddings \cite{glove}.\footnote{The pretrained Glove model we used can be downloaded from http://nlp.stanford.edu/data/glove.42B.300d.zip.}

\vspace{-2mm}
\subsection{Audio-based model}

\begin{figure}[t]
    \centering
    \includegraphics[width=0.90\linewidth]{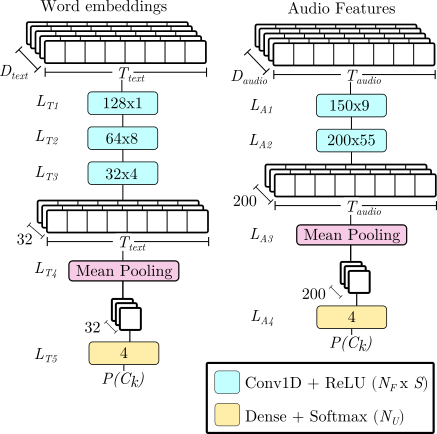}
    \caption{Text-based and audio-based architectures. $T_{text}$ and $T_{audio}$ are the sequence lengths of the model inputs and $D_{text}$ and $D_{audio}$ are the number of features for each input. $N_F$ is the number of convolutional filters, $S$ is the kernel size and $N_U$ is the number of neurons in dense layers. 1D-Convolutional layers operate on the time axis.} 
    \label{fig:branches}
\end{figure}

Each speaker utterance was divided into 32ms segments,  using a hop length of 10ms. The following acoustic features were extracted from each window using openSMILE \cite{opensmile}: pitch, jitter, shimmer, logHNR, loudness, and the first 13 MFCCs. These features were normalized to have a mean of 0 and standard deviation of 1, using the global statistics. Finally, first-order differences were added for all features to form a sequence of 36-dimensional feature vectors that are the input to the neural network shown on the right of Figure~\ref{fig:branches}.

The audio model consists of two convolutional layers $L_{A1}$ and $L_{A2}$ that model the temporal evolution of the input sequence followed by mean-pooling over time. A final dense layer with softmax activation returns the emotion probabilities $P(C_k)$. 

\vspace{-2mm}
\subsection{Fusion models}

In this section, we describe the strategies we implemented to combine audio and text information. In all cases, the fusion model consists of two parallel branches processing audio and text separately up to a layer where the information from the two branches is merged. The models differ on the location of the merging layer, on the network appended after merging, and on the training approach.

\vspace{-2mm}
\subsubsection{Early Fusion}
In the early fusion (EF) approach, the fixed-size embeddings obtained after mean pooling in the audio and text models (Figure~\ref{fig:branches}, layers $L_{T4}$ and $L_{A3}$) are concatenated resulting in a multi-modal embedding of 232 dimensions. This embedding is input to a feed-forward neural network with a hidden dense layer of 128 units with ReLU activation and an output layer with 4 units and softmax activation.

\vspace{-2mm}
\subsubsection{Late Fusion}
In the late fusion model (LF), the logits (pre-softmax) of the audio and text models are concatenated (Figure~\ref{fig:branches}, layers $L_{T5}$ and $L_{A4}$) resulting in an 8-dimensional vector that is used as input to a dense layer of 4 units with softmax activation. This dense layer learns to combine the logits of audio and text modalities to generate the final output probabilities $P(C_k)$. We have also explored learning a scalar weight for each system instead of a full dense layer but the resulting performance was slightly worse.

\vspace{-2mm}
\subsubsection{Training strategies}
We trained our fusion models in 3 different ways: 
 \begin{itemize}[leftmargin=*]
   \setlength\itemsep{0em}
     \item Cold-start (CS): Use Xavier uniform initialization \cite{xavier-init} for all layers of the fusion model and train the model jointly from scratch.
     \item Pre-trained (PT): Train the audio and text models separately and use the trained weights to initialize the corresponding layers of the audio and text branches in the fusion model. The layers after merging are initialized with Xavier uniform initialization. Only these layers are trained, keeping the layers up to the merging point frozen.
     \item Warm-start (WS): Initialize all layers as in the PT approach but instead of training only the layers after merging, train also the layers right before pooling for each branch ($L_{T3}$ and $L_{A2}$), keeping the first layers ($L_{T1}$, $L_{T2}$ and $L_{A1}$) frozen, as in the PT approach. This procedure, in contrast to PT, allows the layers immediately before the pooling to change their weights.
 \end{itemize}

\vspace{-2mm}
\section{Experimental setup and Datasets}
\label{sec:experiments}
\vspace{-2mm}
Our experiments were performed on the IEMOCAP and MSP-PODCAST datasets.
The  Interactive  Emotional  Dyadic  Motion  Capture
(IEMOCAP) dataset \cite{busso-IEMOCAP} has a length of approximately 12 hours and consists of scripted and improvised dialogues by 10 speakers. It is composed of 5 sessions, each including speech from an actor and an actress. Annotators were asked to label each sample choosing one or more labels from a pool of emotions. In this work, we used 4 emotional classes: anger, happiness, sadness and neutral, and following \cite{Fayek2017},  we relabeled excitement samples as happiness. Instances from other classes and with no annotator agreement were discarded.\footnote{Note that discarding no-agreement samples and samples from non-target emotions is not an ideal practice \cite{riera2019}. Here, we decided to do this since it is standard practice in SER literature, facilitating comparisons across papers.} For this dataset, human transcriptions are used for the text-based system.

To test our models we used 5-fold cross-validation, organizing the folds so that training and test sets do not share actors or scripts.
This last point is very important for the text-based model, as has been noted in \cite{gamage2017salience}, because dialogues from the same script are very similar. We show the effect of the criteria used for making the folds on both text and audio models in Section \ref{res:folds}. 

The MSP-PODCAST dataset v1.4 {\cite{busso-podcast}} contains speech segments from podcast recordings, annotated using crowdsourcing. After discarding the instances not belonging to any of the 4 emotional classes under study, the training set contains 12078 speech segments from 601 speakers, while the test set contains 5557 utterances from 50 speakers not present in the training set. The training and test set definitions used in this paper are the ones provided with the dataset. The test set is gender balanced. Speech transcriptions were extracted using the Google Cloud Speech-to-Text API.\footnote{https://cloud.google.com/speech-to-text/}

To counteract the effect of class imbalance present in both datasets, a cost-sensitive training strategy was applied by multiplying the loss of each instance with the inverse of the frequency of the class it belongs to. The models were optimized using Adam \cite{Adam} with a learning rate of 0.0007, except for the fine-tuning case in which the learning rate was decreased to 0.0001, and the case of late fusion using pre-trained branches where learning rate was increased to 0.01. During the first 40 steps, the learning rate was linearly increased from 0 to the final value, except for the late fusion system with pretraining (LF-PT). We applied dropout with 0.5 probability at the input of layer $L_{A2}$ only for the audio branch. As the input sequences have variable length, we padded them with zeros up to a maximum sequence length and then masked the padded values.

We report two different metrics: average recall (AvRec), and the average area under the ROC (AvAUC). Average recall is used instead of accuracy since both datasets have significant imbalance across classes. Both averages are computed over the four target emotions, considering a one-vs-all problem in order to compute individual recall and AUC values. 

We observed that using early stopping in IEMOCAP led to inconsistent results as the data are scarce to generate a validation fold large enough. For this reason, the number of epochs for training each model was selected by optimizing the median AvAUC value on IEMOCAP over 5 seeds. The architectures and hyperparameters were also selected based on IEMOCAP results (sometimes using a single seed).  The final results on IEMOCAP were obtained over 10 seeds, including the 5 used for the optimization of the epoch and the hyperparameters tuning.  This leads to possibly optimistic results on this dataset. On the other hand, the results on MSP-PODCAST were obtained using the same number of epochs and hyperparameters chosen for IEMOCAP, also averaging over 10 seeds. All of our models were trained using Keras \cite{keras}.

\vspace{-2mm}

\section{Results and discussion}
\label{sec:results}
\vspace{-1mm}

In this section we report results for individual and fused systems. We start by showing the effect that the criteria used to define the folds for cross-validation on IEMOCAP has on the two individual systems.
\subsection{Effect of partition criteria for IEMOCAP folds}
\label{res:folds}
\begin{figure}[t]
    \centering
    \includegraphics[width=0.95\columnwidth]{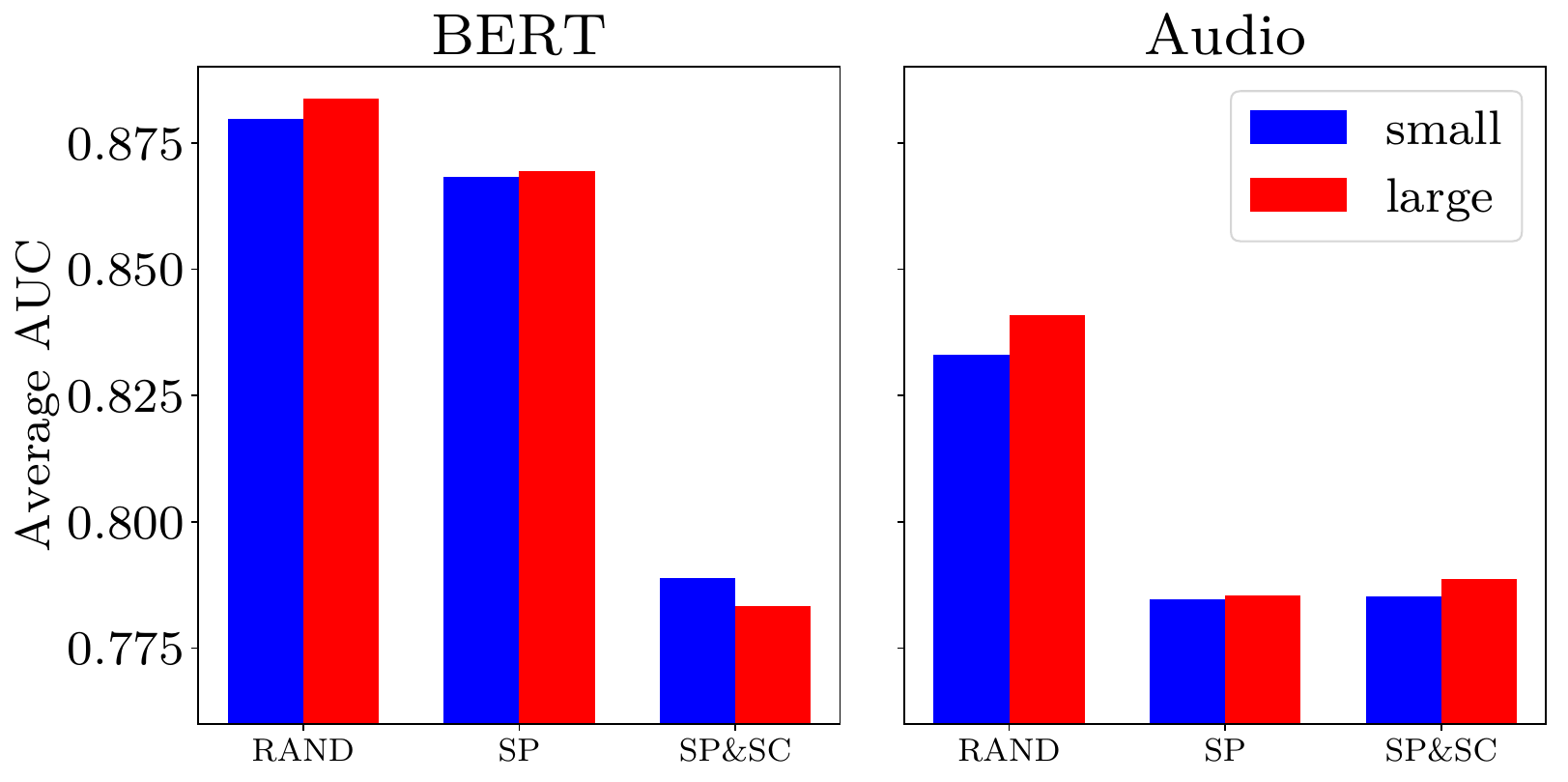}
    \vspace{-2mm}
    \caption{Effect of different criteria for defining the folds in IEMOCAP on audio- and text-based systems for two different model sizes (small and large). RAND: random folds, SP: by-speaker folds, SP\&SC: by-speaker and by-script folds.}
    \label{fig:folds}
\end{figure}

Figure~\ref{fig:folds} shows the AvAUC for three different criteria used to define the folds on IEMOCAP. Results are computed on the merged test scores for all folds. In all cases, 5 folds are used. We compare: (1) random folds (RAND), where no information about speakers or scripts is used to define the folds; (2) folds by speaker (SP) where each fold contains the two speakers from one of the sessions; and (3) fold by speaker and script (SP\&SC) where the folds are defined as in the previous case, but only script 3 is used for testing while all other scripts are used in training. Note that this last option includes less data for each fold. Finally, we compare two different sizes of models: one using half of the nodes in the models from Figure~\ref{fig:branches} (small), and one using twice the number of nodes (large). We note that most papers use by-speaker folds \cite{sebastian_fusion_2019,sahu_multi-modal_2019,xu_learning_2019}, while some use random folds \cite{yoon2018multimodal}. We are not aware of any work that splits by script, though some works discard the scripts altogether using only improvisation instances for testing \cite{gamage2017salience,zhang_exploiting_2019}.

\begin{figure*}[t]
    \centering
    \includegraphics[width=1.0\linewidth]{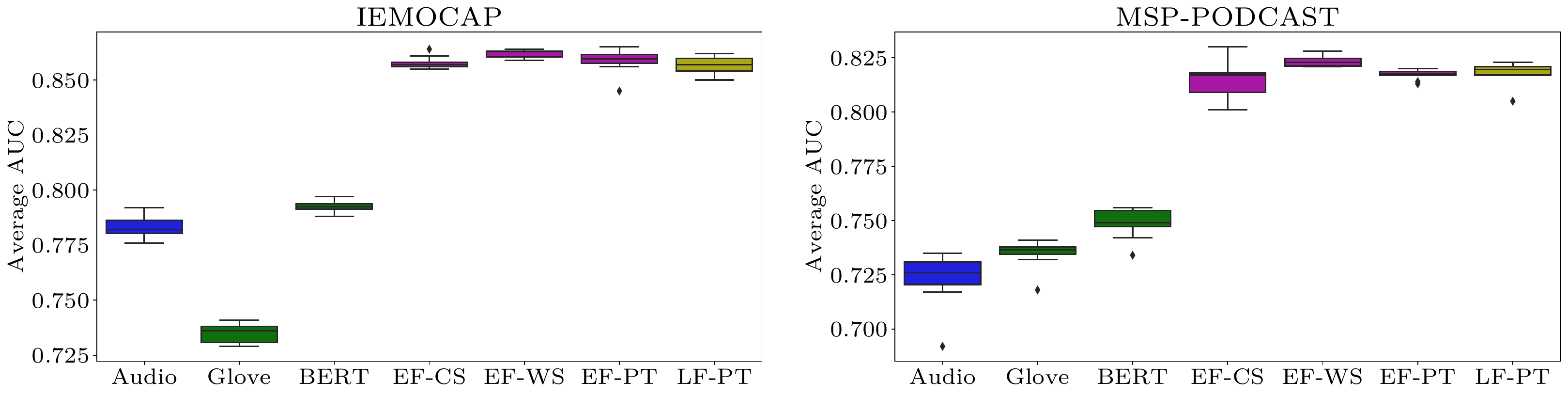}
    \vspace{-6mm}
    \caption{Results for IEMOCAP and MSP-PODCAST dataset. Average AUC distributions for 10 different initialization seeds for different systems: audio model, Glove and BERT based text models, early fusion with cold-start (EF-CS), pretraining (EF-PT) and warm-start (EF-WS) and late fusion with pretraining (LF-PT) models.}
    \label{fig:results}
\end{figure*}

Figure~\ref{fig:folds} clearly shows that both random and by-speaker folds result in optimistic performance for the text-based systems. For the audio-based system, both by-speaker and by-speaker-and-script options lead to similar performance (indicating that the effect of by-speaker-and-script folds having less data is limited), while the random splits result in an optimistic estimation of performance. Furthermore, the conclusion of which  model size is optimal for BERT features changes depending on the fold criteria, as a large model is more likely to overfit, but this effect can only be observed when using folds that do not repeat speakers or scripts between training and test sets. Given these results, we believe it is essential to define the folds for IEMOCAP carefully, not allowing speakers and scripts seen in training to be repeated in testing. In the remaining experiments, we use folds by speaker and script.

\begin{table}
\small
\vspace{1mm}
\caption{Median of evaluation metrics $\pm$ interquartile range obtained using 10 seeds for all the tested models in both datasets.}
\vspace{-1mm}
\label{table:results}
\centering
\begin{tabular}{l|cc|cc} 
\toprule
\multicolumn{1}{c|}{\multirow{2}{*}{ }} & \multicolumn{2}{c|}{IEMOCAP}                                                                  & \multicolumn{2}{c}{MSP-PODCAST}                                                                \\
\multicolumn{1}{c|}{}                   & AvRec (\%)                         & AvAUC                                                   & AvRec (\%)                        & AvAUC                                                     \\ 
\hline
Audio                                   & 56,0$\pm$1,9 & .782$\pm$.006 & 45,7$\pm$1,4 & .726$\pm$.010 \\ 
\hline
Glove                                   & 47,8$\pm$0.1 & .736$\pm$.007 &  49,8$\pm$0.4 & .736$\pm$.003 \\
BERT                                    & 55,2$\pm$1.0 & .792$\pm$.003 &  51,0$\pm$0.9 & .749$\pm$.007 \\ 
\hline
EF-CS                                   & \textbf{65,1$\pm$0.5} & .857$\pm$.002 & 58.2$\pm$2.4 & .817$\pm$.009 \\
EF-WS                                   & 64.7$\pm$1.6 & \textbf{.863$\pm$.002} &  \textbf{59.1$\pm$1.8} & \textbf{.823$\pm$.003} \\ 
EF-PT                                   & 64.9$\pm$1.0 & .859$\pm$.004 & 56.5$\pm$0.3 & .817$\pm$.002 \\
\hline
LF-PT                                   & 63.9$\pm$0.5 & .857$\pm$.006 & 58.0$\pm$0.7 & .819$\pm$.004 \\

\bottomrule
\end{tabular}
\end{table}

\vspace{-2mm}
\subsection{Audio- and text-based models}

Figure~\ref{fig:results} shows the performance obtained with the different systems on both datasets. Average AUC is reported using box and whiskers plots to show the variation in performance for 10 different seeds used to initialize the DNN weights. Table \ref{table:results} shows both AvAUC and AvRec values. We can see that our proposed text model based on BERT embeddings shows  slightly better performance than the audio model on IEMOCAP, while Glove embeddings give significantly worse performance. This contradicts previous results on IEMOCAP, where the text-based models significantly outperform audio models \cite{yoon2018multimodal, sebastian_fusion_2019}. As we showed in Section \ref{res:folds} this is explained by the way we have defined the folds, preventing the text model from being trained in dialogues very similar or identical to the ones present in the test set and avoiding unrealistically good performance estimates for these systems.

The effect of using BERT versus Glove to represent word information can be seen in Figure~\ref{fig:results} and Table \ref{table:results}. BERT embeddings outperform Glove ones in both datasets with relative UAR improvements of 15.5\% and 2.4\% in IEMOCAP and MSP-PODCAST datasets, respectively. We attribute this performance gain to the contextual information imbued in the pretrained BERT model. While our text model could potentially learn contextual information from standard word embeddings like Glove, learning to represent negations or modification values would require a significant amount of data. We hypothesize that this is the reason why Glove performance is closer to BERT in MSP-PODCAST than in IEMOCAP, since the size and variability of dialogues in MSP-PODCAST may be allowing the text model to learn contextual information even from standard word embeddings.

Finally, we note the large effect that the seed has on our systems. In many cases, the ranking of systems changes significantly depending on the seed (results not shown due to lack of space), which thus highlights the critical importance of using several seeds in order to reach more solid conclusions.  

\vspace{-2mm}
\subsection{Fusion models}

As it has been noted in previous works, adding text information to audio-based SER systems gives significant performance improvements \cite{yoon2018multimodal,sahu_multi-modal_2019}. This is also observed in our fusion experiments where for both MSP-PODCAST and IEMOCAP datasets, the AvRec improves 16\% relative to the best performing single model. All fusion approaches perform similarly, in agreement with previous results in the literature  \cite{georgiou_deep_2019, sebastian_fusion_2019}. Only the late fusion approach with pre-training is shown here, due to space considerations. The other two training approaches gave similar results.

A small advantage of the warm-start approach can be observed for both datasets with the early fusion architecture, indicating that this direction may be worth further exploration. In the future, we plan to explore approaches where the fusion is made before or at the pooling layer. We believe this has the potential to give additional benefits since the interaction between both modalities is most likely happening at short time intervals rather than at phrase level.

\vspace{-2mm}
\section{Conclusions}
\label{sec:conclusions}

We presented different approaches for emotion recognition from speech using audio features and transcriptions. We showed results on two publicly available datasets: IEMOCAP and MSP-PODCAST. 
We demonstrated the positive effect of representing linguistic information using contextualized word embeddings extracted with BERT compared to using standard word embeddings like those extracted with Glove. We also showed, in agreement with previous works, that the fusion of audio- and text-based information leads to significant improvements of approximately 16\% on both datasets relative to using the best single modality. To our knowledge, these are the first published results using linguistic information on MSP-PODCAST, a very large, naturalistic and challenging emotion dataset.  

Several fusion strategies were tested, including early and late fusion using different training procedures. Results were not significantly different for the different methods, which again agrees with previous observations in the literature. 

As an additional contribution, we highlighted the importance and impact of how folds are defined for the IEMOCAP dataset, showing how the standard procedure of splitting by session leads to highly optimistic results on our text-based system. We hope that our proposed criteria, which avoids repeating scripted dialogues between training and test sets, or the alternative of discarding scripted dialogues, will be adopted in future works on the IEMOCAP dataset, specially for text-based systems.

\vfill\pagebreak

\bibliographystyle{IEEEbib}
\bibliography{biblio}

\begin{thebibliography}{10}

\bibitem{devillers_challenges_2005}
Laurence Devillers, Laurence Vidrascu, and Lori Lamel,
\newblock ``Challenges in real-life emotion annotation and machine learning based detection,''
\newblock {\em Neural Networks}, vol. 18, no. 4, pp. 407--422, May 2005.

\bibitem{anagnostopoulos_features_2015}
Christos-Nikolaos Anagnostopoulos, Theodoros Iliou, and Ioannis Giannoukos,
\newblock ``Features and classifiers for emotion recognition from speech: a survey from 2000 to 2011,''
\newblock {\em Artificial Intelligence Review}, vol. 43, no. 2, pp. 155--177, Feb. 2015.

\bibitem{zheng_experimental_2015}
W.~Q. Zheng, J.~S. Yu, and Y.~X. Zou,
\newblock ``An experimental study of speech emotion recognition based on deep convolutional neural networks,''
\newblock in {\em 2015 {International} {Conference} on {Affective} {Computing} and {Intelligent} {Interaction} ({ACII})}, Sept. 2015, pp. 827--831.

\bibitem{satt_efficient_2017}
Aharon Satt, Shai Rozenberg, and Ron Hoory,
\newblock ``Efficient {Emotion} {Recognition} from {Speech} {Using} {Deep} {Learning} on {Spectrograms},''
\newblock in {\em Interspeech}, Aug. 2017, pp. 1089--1093.

\bibitem{jin-evector}
Q.~{Jin}, C.~{Li}, S.~{Chen}, and H.~{Wu},
\newblock ``Speech emotion recognition with acoustic and lexical features,''
\newblock in {\em ICASSP}, April 2015, pp. 4749--4753.

\bibitem{chuang2004multi}
Ze-Jing Chuang and Chung-Hsien Wu,
\newblock ``Multi-modal emotion recognition from speech and text,''
\newblock in {\em International Journal of Computational Linguistics \& Chinese Language Processing: Special Issue on New Trends of Speech and Language Processing}, August 2004, vol.~9, pp. 45--62.

\bibitem{gamage2017salience}
Kalani~Wataraka Gamage, Vidhyasaharan Sethu, and Eliathamby Ambikairajah,
\newblock ``Salience based lexical features for emotion recognition,''
\newblock in {\em ICASSP}, 2017, pp. 5830--5834.

\bibitem{softfusion}
B.~{Schuller}, G.~{Rigoll}, and M.~{Lang},
\newblock ``Speech emotion recognition combining acoustic features and linguistic information in a hybrid support vector machine-belief network architecture,''
\newblock in {\em ICASSP}, May 2004, vol.~1, pp. I--577.

\bibitem{haque2019audio}
Albert Haque, Michelle Guo, Prateek Verma, and Li~Fei-Fei,
\newblock ``Audio-linguistic embeddings for spoken sentences,''
\newblock in {\em ICASSP}, 2019, pp. 7355--7359.

\bibitem{yoon2018multimodal}
Seunghyun Yoon, Seokhyun Byun, and Kyomin Jung,
\newblock ``Multimodal speech emotion recognition using audio and text,''
\newblock in {\em 2018 IEEE Spoken Language Technology Workshop (SLT)}, 2018, pp. 112--118.

\bibitem{sebastian_fusion_2019}
Jilt Sebastian and Piero Pierucci,
\newblock ``Fusion {Techniques} for {Utterance}-{Level} {Emotion} {Recognition} {Combining} {Speech} and {Transcripts},''
\newblock in {\em Interspeech 2019}, Sept. 2019, pp. 51--55.

\bibitem{sahu_multi-modal_2019}
Saurabh Sahu, Vikramjit Mitra, Nadee Seneviratne, and Carol Espy-Wilson,
\newblock ``Multi-{Modal} {Learning} for {Speech} {Emotion} {Recognition}: {An} {Analysis} and {Comparison} of {ASR} {Outputs} with {Ground} {Truth} {Transcription},''
\newblock in {\em Interspeech}, Sept. 2019, pp. 3302--3306.

\bibitem{zhang_exploiting_2019}
Biqiao Zhang, Soheil Khorram, and Emily~Mower Provost,
\newblock ``Exploiting {Acoustic} and {Lexical} {Properties} of {Phonemes} to {Recognize} {Valence} from {Speech},''
\newblock in {\em ICASSP}, May 2019, pp. 5871--5875.

\bibitem{devlin2019bert}
Jacob Devlin, Ming-Wei Chang, Kenton Lee, and Kristina Toutanova,
\newblock ``Bert: Pre-training of deep bidirectional transformers for language understanding,''
\newblock in {\em Proceedings of {NAACL-HLT}, Minneapolis, USA}, 2019, pp. 4171--4186.

\bibitem{glove}
Jeffrey Pennington, Richard Socher, and Christopher~D. Manning,
\newblock ``Glove: Global vectors for word representation,''
\newblock in {\em EMNLP}, 2014.

\bibitem{sun_multi-modal_2019}
Zhongkai Sun, Prathusha~K. Sarma, William Sethares, and Erik~P. Bucy,
\newblock ``Multi-{Modal} {Sentiment} {Analysis} {Using} {Deep} {Canonical} {Correlation} {Analysis},''
\newblock in {\em Interspeech}, Sept. 2019, pp. 1323--1327.

\bibitem{busso-IEMOCAP}
Carlos Busso, Murtaza Bulut, Chi-Chun Lee, Abe Kazemzadeh, Emily Mower, Samuel Kim, Jeannette~N. Chang, Sungbok Lee, and Shrikanth~S. Narayanan,
\newblock ``Iemocap: interactive emotional dyadic motion capture database,''
\newblock {\em Language Resources and Evaluation}, vol. 42, no. 4, pp. 335, Nov 2008.

\bibitem{busso-podcast}
Reza Lotfian and Carlos Busso,
\newblock ``Building naturalistic emotionally balanced speech corpus by retrieving emotional speech from existing podcast recordings,''
\newblock {\em IEEE Transactions on Affective Computing}, vol. PP, pp. 1--1, 08 2017.

\bibitem{vaswani2017attention}
Ashish Vaswani, Noam Shazeer, Niki Parmar, Jakob Uszkoreit, Llion Jones, Aidan~N Gomez, {\L}ukasz Kaiser, and Illia Polosukhin,
\newblock ``Attention is all you need,''
\newblock in {\em Advances in neural information processing systems}, 2017, pp. 5998--6008.

\bibitem{opensmile}
Florian Eyben, Martin W\"{o}llmer, and Bj\"{o}rn Schuller,
\newblock ``Opensmile: The munich versatile and fast open-source audio feature extractor,''
\newblock in {\em Proceedings of the 18th ACM International Conference on Multimedia}, New York, NY, USA, 2010, pp. 1459--1462.

\bibitem{xavier-init}
Xavier Glorot and Yoshua Bengio,
\newblock ``Understanding the difficulty of training deep feedforward neural networks,''
\newblock in {\em In Proceedings of the International Conference on Artificial Intelligence and Statistics (AISTATS’10)}, 2010.

\bibitem{Fayek2017}
Haytham~M. Fayek, Margaret Lech, and Lawrence Cavedon,
\newblock ``{Evaluating deep learning architectures for Speech Emotion Recognition},''
\newblock {\em Neural Networks}, vol. 92, pp. 60--68, 2017.

\bibitem{riera2019}
Pablo Riera, Luciana Ferrer, Agust\'in Gravano, and Lara Gauder,
\newblock ``No sample left behind: Towards a comprehensive evaluation of speech emotion recognition system,''
\newblock in {\em Proc. Workshop on Speech, Music and Mind 2019}, 2019.

\bibitem{Adam}
Diederik~P. Kingma and Jimmy Ba,
\newblock ``Adam: A method for stochastic optimization,''
\newblock in {\em 3rd International Conference for Learning Representations}, 2015.

\bibitem{keras}
Fran\c{c}ois Chollet et~al.,
\newblock ``Keras,'' https://keras.io, 2015.

\bibitem{xu_learning_2019}
Haiyang Xu, Hui Zhang, Kun Han, Yun Wang, Yiping Peng, and Xiangang Li,
\newblock ``Learning alignment for multimodal emotion recognition from speech,''
\newblock in {\em Interspeech}, 2019, pp. 3569--3573.

\bibitem{georgiou_deep_2019}
Efthymios Georgiou, Charilaos Papaioannou, and Alexandros Potamianos,
\newblock ``Deep {Hierarchical} {Fusion} with {Application} in {Sentiment} {Analysis},''
\newblock in {\em Interspeech 2019}, Sept. 2019, pp. 1646--1650.

\end{thebibliography}

\end{document}